\documentclass[10pt]{article}

\usepackage[margin=1in]{geometry}

\usepackage[utf8]{inputenc}
\usepackage[T1]{fontenc}
\usepackage[english]{babel}

\usepackage{amsmath, amssymb, amsfonts, amsthm, mathtools, bm}
\usepackage{siunitx}

\usepackage{graphicx}
\usepackage{float}
\usepackage{tikz}
\usepackage{tikz}
\usepackage{pgfmath}
\usetikzlibrary{babel}
\usetikzlibrary{intersections}
\usetikzlibrary{fadings}
\usetikzlibrary{calc}
\usetikzlibrary{math}
\usetikzlibrary{fit}
\usetikzlibrary{positioning}
\usetikzlibrary{patterns}
\usetikzlibrary{shapes, arrows, arrows.meta}
\usetikzlibrary{shapes.geometric, shapes.multipart}
\usetikzlibrary{decorations.pathreplacing}
\usetikzlibrary{decorations.pathmorphing}
\usetikzlibrary{decorations.text}
\usetikzlibrary{backgrounds}
\pgfdeclarelayer{background}
\pgfdeclarelayer{foreground}
\pgfsetlayers{background,main,foreground}

\usepackage{booktabs}
\usepackage{multirow}
\usepackage{tabularx}
\newcolumntype{C}{>{\centering\arraybackslash}X}
\newcolumntype{L}{>{\raggedright\arraybackslash}X}

\usepackage{algorithmic}
\usepackage{listings}

\usepackage[hidelinks]{hyperref}
\usepackage{url}

\usepackage{xcolor}
\usepackage[normalem]{ulem}

\newcommand{\vect}[1]{\bm{#1}}
\newcommand{\matr}[1]{\bm{#1}}
\newcommand{\quat}[1]{\overline{\vect{#1}}}
\newcommand{\Dquat}[1]{\hat{\vect{#1}}}
\newcommand{\Tquat}[1]{\breve{\vect{#1}}}

\newcommand{\I}{\matr{I}}
\newcommand{\0}{\matr{0}}

\newcommand{\PrincipalPart}{\mathcal{P}}
\newcommand{\DualPart}{\mathcal{D}}

\newcommand{\qbe}{\quat{q}_{b}^{e}}

\newcommand{\wibb}{\vect{\omega_{ib}^{b}}}
\newcommand{\Dwibb}{\vect{\delta\omega_{ib}^{b}}}
\newcommand{\Tqbe}{\Tquat{q}_{b}^{e}}
\newcommand{\TqEst}{\tilde{\Tquat{q}}}
\newcommand{\TqbeEst}{\tilde{\Tquat{q}}_{b}^{e}}
\newcommand{\TqbeEstconj}{\tilde{\Tquat{q}}_{b}^{e*}}
\newcommand{\qbeEst}{\tilde{\quat{q}}_{b}^{e}}
\newcommand{\qbeEstconj}{\tilde{\quat{q}}_{b}^{e*}}

\newcommand{\qprod}{.}

\title{\textbf{Enhanced UAV Navigation Systems through Sensor Fusion with Trident Quaternions}}

\author{
Sebastian Incicco\textsuperscript{1}, 
Juan Ignacio Giribet\textsuperscript{2}, 
Leonardo Colombo\textsuperscript{3} \\
\\
\textsuperscript{1}Facultad de Ingeniería, Universidad de Buenos Aires, Argentina \\
\textsuperscript{2}Universidad de San Andrés and CONICET, Argentina \\
\textsuperscript{3}Centro de Automática y Robótica (CAR, UPM-CSIC), Madrid, España
}

\date{}

\begin{document}

\maketitle

\begin{abstract}
This paper presents an integrated navigation algorithm based on trident quaternions, an extension of dual quaternions. The proposed methodology provides an efficient approach for achieving precise and robust navigation by leveraging the advantages of trident quaternions. The performance of the navigation system was validated through experimental tests using a multi-rotor UAV equipped with two navigation computers: one executing the proposed algorithm and the other running a commercial autopilot, which was used as a reference.
\end{abstract}


\section{Introduction} 
\label{sec:Intro}

In the field of navigation systems, achieving accurate and reliable estimates of a vehicle's position, velocity, and orientation remains a challenge. Integrated Navigation (IN) techniques have emerged as a promising solution by combining multiple sensor measurements, such as those obtained from Inertial Measurement Units (IMU), Global Navigation Satellite Systems (GNSS), and vision-based sensors. IN approaches offer significant advantages, including robustness, improved accuracy, and the ability to overcome the limitations of individual sensors.

Among the various mathematical tools employed in IN, quaternions have garnered considerable attention for estimating a vehicle’s attitude (orientation). Quaternions provide an elegant and compact representation of orientation, avoiding the limitations of traditional Euler angles, such as singularities and ambiguity. Moreover, the use of quaternions facilitates seamless integration of sensor measurements, enabling efficient data fusion and enhanced navigation performance. On the other hand, dual quaternions are mathematical constructs that have proven useful in various domains, such as computer vision, geometric representation, and robotics\cite{Lynch2017}. In mobile robotics, for instance, they are beneficial due to their concise and intuitive representation of both position and attitude within a single algebraic equation.

In \cite{Fike2011} the authors introduce the concept of hyperdual numbers to enable precise second-derivative computations avoiding truncation, this idea was extended later to accurately model multibody kinematics in a compact form. Following the ideas of \cite{Fike2011}, in \cite{TQ1} and \cite{TQ2} the authors develop a compact and efficient representation of the strap-down Inertial Navigation System (INS). For this purpose,  an extension of dual quaternions, known as trident quaternions, was proposed. These allow for the compact representation of a vehicle's navigation variables (velocity, position, and orientation) within a single mathematical object, preserving the fundamental properties of dual quaternions. Like quaternions and dual quaternions, trident quaternions share a similar mathematical treatment, as they are specific instances of what is known as Clifford algebras \cite{TQ3}.  

Following the ideas of \cite{TQ1}, \cite{TQ2} and \cite{TQ3}, this paper describes the development and implementation of a navigation system that combines an inertial navigation algorithm (INS) with an Extended Kalman Filter (EKF), utilizing trident quaternions. The formulation of navigation equations in terms of trident quaternions provides certain benefits. Similar to traditional quaternions used for attitude representation, trident quaternions enable a numerically stable formulation of algorithms through normalization. This is crucial for preventing issues such as the propagation of numerical errors. Additionally, they have the capacity to encapsulate both rotational and translational movements within a single equation, akin to dual quaternions.

However, what distinguishes trident quaternions is their ability to include the velocity information of the vehicle. This inclusion is critical for appropriately formulating inertial navigation equations given that measurements provided by the IMU contain information about acceleration and angular velocity, but not linear velocity. Explicitly incorporating vehicle velocity into the equations enhances the accuracy and robustness of the navigation system.

While it could be argued that there is no clear advantage in terms of efficiency or performance when comparing the dual quaternion-based implementation to this new trident quaternion-based representation, there is undoubtedly an improvement in the simplicity of the navigation equations. When properly handled, this simplicity translates into a more straightforward software implementation—not only for embedded hardware but also for testing techniques applied to the code. This is because the navigation equations adopt the same form as the attitude equations described by quaternions, which, in turn, are the same equations that dual quaternions use to represent both position and orientation.

The remainder of the paper is as follows. Section \ref{sec:Intro} presents known results on quaternions, dual quaternions, and trident quaternions, which will be instrumental in the rest of the work. Section \ref{sec:Navega} presents the navigation equations formulated using a trident quaternion-based framework. Section \ref{sec:Kalman} introduces the equations used to fuse inertial navigation data with information from external sensors, in this case, GPS. Section \ref{sec:Results} presents experimental tests conducted on a multi-rotor UAV, demonstrating that the navigation results are consistent when compared to information obtained from a reference autopilot. Finally, Section \ref{sec:Conclusions} provides the conclusions of this work.

\subsection{Quaternions} 

Let $\vect{p}\in\mathbb{R}^3$ represent the vehicle's position, and let $a$ denote a reference frame. Then, $\vect{p}^a$ denotes the vehicle's position expressed in frame $a$. 

Quaternions can be thought of as a four-dimensional generalization of the complex numbers defined as:
\begin{align*}
\bf{H} := &\{ \overline{q} = q_0 + q_1i + q_2j + q_3k \mid q_i \in\mathbb{R},\\
&\, i^2 = j^2 = k^2 = ijk = -1 \},    
\end{align*}
\noindent where multiplication is non-commutative. Every quaternion $\overline{q}\in \bf{H}$ has a real and imaginary part, denoted as $q_0$ and $\vect{q}$, respectively.s


Every quaternion $\quat{q}\in \bf{H}$ has a real and imaginary part, denoted as $q_0$ and $\vect{q}$, respectively.
%
Analogously to the complex numbers, its conjugate can be defined for \( \quat{q} = q_0 + q_1i + q_2j + q_3k  \):
\[
\quat{q}^* = q_0 - q_1i - q_2j - q_3k,
\]
so that \(\quat{q}\qprod\quat{p} = \quat{p}^* \qprod \quat{q}^*\), and its norm as
\(
\|\quat{q}\| = \sqrt{\quat{q} \qprod \quat{q}^*}
\).
The inverse of \( \quat{q} \neq 0 \) is $\quat{q}^{-1} = {\quat{q}^*}/{\|\quat{q}\|^2}$.

In particular, the set of  unit quaternions $\bf{H}_1$ is widely used in robotics being a consistent way to represent attitude. 
Unit quaternions are a Lie group under multiplication.
The inverse of multiplication is reduced to conjugation and the unit is the real quaternion $1$. To each element in $\bf{H}_1$ it is possible to assign an element in $SO(3)$:
\[
\matr{C} : \bf{H}_1 \to SO(3), \quad \quat{q} \mapsto R(\quat{q}) = I_3 + 2q_0 (\vect{q}\times) + 2 (\vect{q}\times)^2,
\]
where \( (\cdot)\times : \mathbb{R}^3 \to \text{so}(3) \) is the standard map of vectors to skew-symmetric \( 3 \times 3 \) matrices. This formula is a two to one map given that $\matr{C}(\quat{q})=\matr{C}(-\quat{q})$.

The Lie algebra of unit quaternions is the set of purely imaginary quaternions, i.e., $\bf{h_1} = \left\{ \quat{q} \in \bf{H} : \quat{q} = 0 + {q} \right\}$. Every $\quat{q}\in \bf{h_1}$ is
univocally related with one element $q\in\mathbb{R}^3$. 
Given $x\in\mathbb{R}^3$, we denote $\quat{x}\in \bf{h_1}$ as the quaternion with zero real part, and imaginary part equal to $x$. In a similar way, for \(\quat{q}\in \bf{h_1}\), the inverse of this map is denoted as ${q}\in\mathbb{R}^3$. 

Given $\overline{q}=q_0+\vect{q}$, the exponential and logarithm can be defined in terms of power series:
\[\exp(\quat{q}) = e^{q_0} \left( \cos(\|\vect{q}\|) + \frac{\sin(\|\vect{q}\|)}{\|\vect{q}\|} \vect{q} \right),\]
\[\ln(\quat{q}) = \ln(\|\quat{q}\|) + \cos^{-1} \left(\frac{q_0}{\|\vect{q}\|} \right) \frac{\vect{q}}{\|\vect{q}\|}.\]

Also, given $x\in\mathbb{R}^3$, it is possible to assign an element in $\bf{H}_1$, as: 
\(
exp(x) = \cos(\|{x}\|) + \frac{\sin(\|{x}\|)}{\|{x}\|} {x}
= \cos\left({\theta}/{2}\right) + \sin\left({\theta}/{2}\right) {n},
\)
where ${n}\in\mathbb{R}^3$ is a unit vector representing the axis of rotation and $\theta$ is the angle of rotation. 

For quaternions $\quat{v}\in \bf{h_1}$ and $\quat{q}\in \bf{H}_1$, the Adjoint transformation is defined as \(Ad_{\quat{q}} \quat{v} = \quat{q} \qprod \quat{v} \qprod \quat{q}^*\).

The set $\bf{H}$ can be identified with $\mathbb{R}^4$ and its operations can be expressed in matrix form.
Thus, for $\quat{p},\quat{q}\in\bf{H}$, with $\quat{p}=(\vect{p},p_0)$ and $\quat{q}=(\vect{q},q_0)$, in matrix form the quaternion product $\qprod$ can be written as:
\begin{equation}\label{quat_prod}
    \quat{p}\qprod\quat{q} =
        \begin{bmatrix}
            (\vect{p})\times + \I p_0 & \vect{p} \\
                -\vect{p}^T      &    p_0
        \end{bmatrix}
        \begin{pmatrix}
            \vect{q} \\
            q_0
        \end{pmatrix},
\end{equation}
where $\I\in\mathbb{R}^{3\times 3}$ is the identity matrix. 

Suppose $\qbe=(\vect{q_b^e},q_{b_{0}}^e)\in\bf{H}_1$ is a unit quaternion representing the rotation between the vehicle frame $b$ and the ECEF frame $e$ ({\it Earth-Centered Earth-Fixed}), then $\vect{p}^b$ and $\vect{p}^e$ are related in terms of quaternions as $\quat{p}^e = Ad_{\qbe} \quat{p}^b$, where $\quat{p}^b=(\vect{p}^b,0)$ and $\quat{p}^e=(\vect{p}^e,0)\in \bf{h_1}$. Additionally, let $\displaystyle{\matr{C}(\qbe)} = \matr{C_b^e}$ such that $\vect{p}^e = \matr{C_b^e}  \vect{p}^b$. Let $\vect{\omega}=\vect{\omega}_{eb}^b (t)\in\mathbb{R}^3$ denote the angular velocity of frame $b$ relative to $e$, expressed in $b$, and let $\quat{\omega}=(\vect{\omega},0)$. Then, the dynamics of the unit quaternion $\qbe$ is given by
\vspace{-2 mm} 
\begin{equation}\label{quat_dynamic}
    \dot{\qbe}  = \frac{1}{2} \qbe \qprod \quat{\omega}%
                    = \frac{1}{2}
                        \begin{bmatrix}
                            (\vect{q_b^e})_\times + \I q_{b_{0}}^e \\
                            -\vect{q_b^{e^T}}
                        \end{bmatrix}\vect{\omega}.
\end{equation}

\subsection{Dual Quaternions} 

A dual number $\Dquat{\alpha}$ is defined as $\Dquat{\alpha}=a+\varepsilon b$, where $a$ and $b$ are real numbers, referred to as the real part and dual part, respectively, with $\varepsilon$ being an element such that $\varepsilon\neq0$ and $\varepsilon^2=0$. This concept extends to dual quaternions, which have proven useful in fields such as computer graphics and robotics, particularly for representing the position and attitude of a vehicle in a single mathematical object~\cite{Lynch2017, adorno2017robot}.

The set of dual quaternions can be defined as:
\begin{align*}
\bf{D} = \left\{ \hat{\vect{q}} = \quat{q}_P + \varepsilon \quat{q}_D  \mid \quat{q}_P, \quat{q}_D \in \bf{H} \right\}.
\end{align*}
\noindent where $\PrincipalPart(\Dquat{q}) = \quat{q}_P$ is the principal part, and $\DualPart(\Dquat{q}) = \quat{q}_D$ its dual part.
Conjugation is naturally extended for any dual quaternion \( \hat{\vect{q}} = \quat{q}_P + \varepsilon\quat{q}_D  \) as \( {\hat{\vect{q}}^*} = \quat{q}_P^* + \varepsilon\quat{q}_D^*  \). So is the norm, $\|\hat{\vect{q}}\| = \sqrt{{\hat{\vect{q}}}^* {\hat{\vect{q}}}}$.
Unit dual quaternions are defined as:
\[
\bf{D}_1 = \left\{ \hat{\vect{q}} \in \mathcal{D} \mid \|\hat{\vect{q}}\| = 1 \right\},
\]
they are Lie group with the inverse of a unit dual quaternion being its conjugate. Every unit dual quaternion can be written as
\(
\hat{\vect{q}}= \quat{q} + \varepsilon\frac{1}{2} \quat{p} \, \quat{q},
\)
\noindent where $\quat{q}\in \bf{H}_1$ is a unit quaternion representing a rotation and $\quat{p}\in \bf{h_1}$ is purely imaginary quaternion representing a translation.


The Lie algebra of \( \bf{D}_1 \) is the set of purely imaginary dual quaternions, i.e.,
\[\bf{d}_1 := \left\{ \hat{x} \in \bf{D} \mid {\hat{x}^*} + \hat{x} = 0 \right\} \cong \mathbb{R}^3 \times \mathbb{R}^3.\]


The addition, product, and conjugation of dual quaternions can be extended from $\bf{H}$, considering that $\varepsilon^2=0$. The conjugate of a dual quaternion is given by $\Dquat{q}^* = \PrincipalPart(\Dquat{q})^* + \varepsilon \DualPart(\Dquat{q})^*$. The identity element of the set of dual quaternions, denoted by ${\Dquat{I}}$, is defined as ${\Dquat{I}} = \quat{I} + \varepsilon \quat{0}$, where $\quat{0}=(\vect{0},0)$, such that $\Dquat{q} \qprod {\Dquat{I}} = {\Dquat{I}} \qprod \Dquat{q} = \Dquat{q}$.

Given $\Dquat{q}$, the attitude and position of the vehicle can be recovered as follows: $\quat{q} = \PrincipalPart(\Dquat{q})$ and $\quat{p} = 2\DualPart(\Dquat{q}) \qprod \PrincipalPart{(\Dquat{q})}^*$. The time derivative of $\Dquat{q}$ can be obtained from the derivatives of its principal and dual parts. Specifically,
\begin{equation}\label{quat-dual-dynamic}
\dot{\PrincipalPart}(\Dquat{q}) = \dot{\quat{q}}
                              = \frac{1}{2}\quat{q}\qprod\quat{\omega},\,\,
    2\dot{\DualPart}(\Dquat{q}) = \DualPart(\Dquat{q}) \qprod
        \quat{\omega} + \quat{v}\qprod\PrincipalPart(\Dquat{q}),
\end{equation} 
where $\quat{v}=\dot{\quat{p}}$, and the last equality follows from equation~\eqref{quat_dynamic}. The temporal evolution of $\Dquat{q}$, and hence the vehicle's pose, is governed by the commanded angular velocity $\vect{\omega}$ (expressed in the vehicle frame) and the linear velocity $\vect{v}$ (expressed in frame $e$).

\subsection{Trident Quaternions}

The concept of dual numbers can be extended to define a trident number $\Tquat{\alpha}$ as $\Tquat{\alpha}=a+\varepsilon_1 b_1 + \varepsilon_2 b_2$, where $a$, $b_1$, and $b_2$ are real numbers, with $\varepsilon_1, \varepsilon_2 \neq 0$ and $\varepsilon_1^2=\varepsilon_2^2=\varepsilon_1\varepsilon_2=0$.

Similarly, the trident quaternion is an extension of the dual quaternion. Given a unit quaternion $\quat{q} \in \bf{H}_1$ (representing the vehicle's attitude), $\quat{v} \in \bf{h_1}$ (representing the vehicle's velocity), and $\quat{p} \in \bf{h_1}$ (representing the vehicle's position), unitary trident quaternions are defined as $\Tquat{q} = \quat{q} + \varepsilon_1 \frac{1}{2} (\quat{v}\qprod\quat{q}) + \varepsilon_2 \frac{1}{2} (\quat{p}\qprod\quat{q})$, where $\PrincipalPart(\Tquat{q}) = \quat{q}$ is the principal part, and $\DualPart_1(\Tquat{q}) = \frac{1}{2}(\quat{v}\qprod \quat{q})$ and $\DualPart_2(\Tquat{q}) = \frac{1}{2}(\quat{p}\qprod \quat{q})$ are the primary and secondary imaginary parts of $\Tquat{q}$, respectively. The conjugate of the trident quaternion is given by $\Tquat{q}^* = \PrincipalPart(\Tquat{q})^* + \varepsilon_1 \DualPart_1(\Tquat{q})^* + \varepsilon_2 \DualPart_2(\Tquat{q})^*$. Trident quaternions with zero real components in their principal part and imaginary parts will be referred to as {\it trident quaternion vectors}.

Observe that given $\Tquat{q}$, it is possible to recover the vehicle's attitude, velocity, and position as follows: $\quat{q} = \PrincipalPart(\Tquat{q})$, $\quat{v} = 2\DualPart_1(\Tquat{q}) \qprod \PrincipalPart{(\Tquat{q})}^*$, and $\quat{p} = 2\DualPart_2(\Tquat{q}) \qprod \PrincipalPart{(\Tquat{q})}^*$. The time derivative of $\Tquat{q}$ can be obtained from the derivatives of the principal and imaginary parts. Therefore
\begin{align}\label{quat-trid-dynamic}
\dot{\PrincipalPart}(\Tquat{q}) &= \dot{\quat{q}}
                              = \frac{1}{2}\quat{q}\qprod\quat{\omega}, \nonumber \\
    2\dot{\DualPart_1}(\Tquat{q}) &= \DualPart_1(\Tquat{q}) \qprod
        \quat{\omega} + \quat{a}\qprod\PrincipalPart(\Tquat{q}) \nonumber \\
    2\dot{\DualPart_2}(\Tquat{q}) &= \DualPart_2(\Tquat{q}) \qprod
        \quat{\omega} + \quat{v}\qprod\PrincipalPart(\Tquat{q})
\end{align} where $\quat{a}=\dot{\quat{v}}=\Ddot{\quat{p}}$. The temporal evolution of $\Tquat{q}$ is determined by the commanded angular velocity $\vect{\omega}$ (in the vehicle frame), the acceleration $\vect{a}$, and the linear velocity $\vect{v}$ (both in the inertial frame $e$).  
Let ${\Tquat{I}}$ be the identity element of the set of trident quaternions, defined as ${\Tquat{I}} = \quat{I} + \varepsilon_1 \quat{0} + \varepsilon_2 \quat{0}$, such that $\Tquat{q} \qprod {\Tquat{I}} = {\Tquat{I}} \qprod \Tquat{q} = \Tquat{q}$.

\section{Navigation Equations}
\label{sec:Navega}
In the field of navigation systems, the so-called navigation equations are equations that relate the position, velocity, and attitude of a vehicle to its angular velocity and specific force. This is because inertial navigation systems use an IMU to measure the specific force and angular velocity of a vehicle. These equations allow for the determination of the so-called navigation variables: position, velocity, and orientation, which are essential for tracking and controlling the movement of a vehicle in a specific environment, whether on the Earth's surface, in the air, at sea, or in space. In particular, this work uses a {\it strapdown} configuration, which means that the IMU sensors are fixed to the vehicle, so their measurements are expressed in the vehicle's frame of reference \cite{Rogers,Espana}.

The classical navigation equations expressed between the ECEF frame $e$ and the body frame $b$, which, given the initial conditions, the inertial system is responsible for solving to obtain the navigation variables, are as follows
\begin{align}
\dot{\vect{p}}^e &= \vect{v}^e,\\
\dot{\vect{v}}^e &= \matr{C_b^e} \vect{f}^b-2(\vect{\omega}_{ie}^e)_\times\vect{v}^e-\left((\vect{\omega}_{ie}^e)_\times\right)^2\vect{p^e}+\vect{\tilde{\gamma}}^e (\vect{p^e} ),\\
\dot{\quat{q}}_{b}^{e} &=\frac{1}{2} \quat{q}_{b}^{e} \qprod \quat{\omega}_{eb}^b =\frac{1}{2} \quat{q}_{b}^{e} \qprod \quat{\omega}_{ib}^b - \frac{1}{2} \quat{\omega}_{ie}^e \qprod \quat{q}_{b}^{e}, 
\end{align}
\noindent where $\vect{f^b}, \vect{\omega_{ib}^b}$ are the specific forca and angular velocity, which are the inertial measurements provided by the IMU,  $\vect{\omega_{ie}^e}$ is the Earth's rotation rate and $\vect{\tilde{\gamma^e}}(\vect{p^e})$ is a model of gravity, which depends on the vehicle's position. In what follows $\vect{{\gamma^e}}(\vect{p^e})=\vect{\tilde{\gamma^e}}(\vect{p^e})-\left((\vect{\omega}_{ie}^e)_\times\right)^2\vect{p^e}$. Additionally, $\quat{\omega}_{ie}^e=(\vect{\omega_{ie}^e},0)$ and $\quat{\omega}_{ib}^b=(\vect{\omega_{ib}^b},0)$. 

One of the main drawbacks of inertial navigation systems is that their error can grow unbounded over time \cite{Rogers}. This error mainly depends on the sensor's characteristics, the numerical integration algorithm, and the knowledge of the vehicle's initial conditions. It's important to understand how the inertial navigation error evolves over time, which can be obtained by performing a first-order analysis of its dynamics. For this, it is necessary to introduce a model for the measurements of the inertial instruments. A commonly used model when working with MEMS sensors is the following:
\begin{equation}
\vect{\wibb} = \vect{\tilde{\omega}_{ib}^b} + \vect{b_{\omega}} + \vect{\xi_{\omega}}, 
\vect{f_{b}} = \vect{\tilde{f}_{b}} + \vect{b_{f}} + \vect{\xi_{f}}, 
\end{equation} where $\vect{\wibb}$ and $\vect{f_{b}}$ are the true values, while $\vect{\tilde{\omega}_{ib}^b}$ and $\vect{\tilde{f}_{b}}$ are the readings of the inertial sensors. On the other hand, $\vect{b_{\omega}} (\vect{b_{f}})$ and $\vect{\xi_{\omega}} (\vect{\xi_{f}})$ represent the biases and noise in the measurements, respectively, in the gyroscopes (accelerometers). These biases are considered random variables subject to variations. A common criterion to estimate the variability of these parameters is to consider them as stochastic processes and include them as state variables in a data fusion filter to correct the inertial measurements in real-time. For example, in the EKF model presented in the next section, the instabilities of the biases $\vect{\delta b_f}$ and $\vect{\delta b_{\omega}}$ are included in the error state vector. Therefore, the error model for the inertial measurements is as follows:
\begin{align}
\vect{\delta \wibb} &= \vect{\delta b_{\omega}} + \vect{\xi_{\omega}}, \vect{\quad \delta \dot{b}_{\omega}} = \vect{\xi_{b_{\omega}}} \label{eq_Domega}, \\ 
\vect{\delta f_{b}} &= \vect{\delta b_{f}} + \vect{\xi_{f}}, \vect{\quad \delta \dot{b}_{f}} = \vect{\xi_{b_{f}}}, \quad \label{eq_Dfb}
\end{align}

\noindent where $\vect{\Dwibb}$ and $\vect{\delta f_{b}}$ are the measurement errors, while $\vect{\delta\dot{b}_{\omega}}$ and $\vect{\delta\dot{b}_{f}}$ represent the temporal evolution of the bias instabilities, which are assumed to be white Gaussian noise with zero mean and known variance.
The dynamics of the nonlinear system error model are

\begin{align}
\vect{\delta\dot{p}^e} &= \vect{\delta{v}^e}, \label{eq_pe} \\
\vect{\delta\dot{v}^e} &= -(\matr{C_b^e} \vect{f^b})_\times\vect{\phi^e} + \matr{C_b^e}\vect{\delta f^b} + \left(\frac{\partial \vect{\gamma^e}}{\partial \vect{p^e}}\right) \vect{\delta p^e} - \\ &2(\vect{\Omega_e^e})_\times\vect{\delta v^e},\label{eq_ve} \\
\quat{\delta q}_b^b &= (\sin(\|\vect{\phi^b}\|/2)\vect{\phi^b},\cos(\|\vect{\phi^b}\|/2)),\\
\vect{\dot{\phi}^b} &=
-(\wibb)_\times \vect{\phi^b} - \vect{\Dwibb}, \label{eq_phib}  
\end{align}
\noindent where $\vect{\phi^b}=ln(\qbe)$ is the angular error vector associated with the quaternion $\qbe\in\bf{H}_1$ and \(\left(\frac{\partial \vect{\gamma^e}}{\partial \vect{p^e}}\right)\) is the Jacobian of the gravity model used, which depends on the vehicle's current position. Here $\|.\|$ represents the 2-norm. The state variable modeling the inertial system's error is given by
\begin{align}
    \vect{\delta x}=\begin{bmatrix}
        \vect{\phi^b} & \vect{\delta p^e} & \vect{\delta v^e} & \vect{\delta b_{\omega}} & \vect{\delta b_f}
    \end{bmatrix},\label{eq:INSerror}
\end{align}
\noindent where its mean $E(\vect{\delta x}(0))$ and covariance $\matr{P_{\delta x}}(0)>0$ at the initial time are considered known. Thus, by using information from $\vect{\delta x}$, for example, provided by an external sensor, it can be estimated using a Kalman filter \cite{Espana}.

It is possible to represent a vehicle's navigation variables compactly by defining the unit trident quaternion $\Tquat{q}_{b}^{e}$:
\begin{equation}\label{eq:INS-tridente}
\Tquat{q}_{b}^{e} = \quat{q}_{b}^{e} + \varepsilon_1 \frac{1}{2} \quat{v}^{e} \qprod \quat{q}_{b}^{e} + \varepsilon_2 \frac{1}{2} \quat{p}^{e} \qprod \quat{q}_{b}^{e}.
\end{equation}

Moreover, by analyzing the time evolution of $\Tquat{q}_{b}^{e}$, the navigation equations can be formulated in terms of trident quaternions as:
\begin{equation}
\dot{\Tquat{q}}_{b}^{e} = 
 \frac{1}{2} \Tquat{q}_{b}^{e} \qprod \Tquat{\omega}_{ib}^b - \frac{1}{2} \Tquat{\omega}_{ie}^e \qprod \Tquat{q}_{b}^{e}, 
\end{equation}
\noindent where, 
\small
\begin{align}
\Tquat{\omega}_{ib}^b &= \quat{\omega}_{ib}^b + \varepsilon_1 \quat{f}^{b} \\ \footnotesize
\Tquat{\omega}_{ie}^e &= \quat{\omega}_{ie}^e - \varepsilon_1 \left( \quat{g}^e - \quat{\omega_{ie}^e \times {v}^{e}} \right) - \varepsilon_2 \left( \quat{v}_{e} + \quat{\omega_{ie}^e \times {p}^{e}} \right). 
\end{align}
\normalsize

This result is significant because it extends the expression of the kinematics of a conventional unit quaternion. Note that $\Tquat{\omega}_{eb}^b$, $\Tquat{\omega}_{ib}^b$, and $\Tquat{\omega}_{ie}^e$ are trident quaternions with null real parts, and in fact, $\quat{\omega_{ie}^e \times{p}^{e}}=((\vect{\omega}_{ie}^e)_\times \vect{p}^{e},0)$ and $\quat{\omega_{ie}^e \times{v}^{e}}=((\vect{\omega}_{ie}^e)_\times \vect{v}^{e},0)$.

\section{Extended Kalman Filter} \label{sec:Kalman}

An inertial navigation scheme is not suitable when the IMUs are of low quality. Particularly when MEMS sensors are used, it is necessary to include additional navigation information. External sensors, such as GNSS, magnetometers, barometers, LIDAR, cameras, radars, and ultrasonic sensors, can provide complementary information that helps to improve navigation accuracy and reliability.

In addition to improving accuracy, the inclusion of external sensors can also enhance the robustness of the integrated navigation system by providing data redundancy. By integrating data from external sensors, the system can more effectively detect and correct errors in the inertial sensors, resulting in more reliable and accurate navigation even in challenging conditions. The integration of information from different sensors is done through a data fusion filter, typically an Extended Kalman Filter (EKF).

Let $\qbeEst$ be the estimated quaternion, and let $\qbe$ be the true quaternion. We can define the errors, in terms of quaternions, by left and right as follows: 

\begin{align} \quat{\delta q}_{L} &= \qbeEstconj \qprod \qbe \approx \quat{I} + \frac{1}{2}\quat{\delta \sigma}L \label{delta_q_Left} \\ \quat{\delta q}_{R} &= \qbe \qprod \qbeEstconj \approx \quat{I} + \frac{1}{2}\quat{\delta \sigma}_r \label{delta_q_rigth} \end{align}

Where $\quat{\delta \sigma}_L=(\vect{\delta{\sigma}_L},0)$ and $\quat{\delta \sigma}_r = (\vect{\delta{\sigma}_r},0)$ are quaternion vectors. $\quat{\delta \sigma}_L$ can be understood as the attitude estimation error in terms of the body, while $\quat{\delta \sigma}_r$ is with respect to the error in the earth frame. Similarly, the error definition can be extended to the case of trident quaternions.

\subsection{Error Model of the Left-Handed Trident Quaternion}

The extension to trident quaternions of the expression \eqref{delta_q_Left} can be expressed as follows:
\small
\begin{align}
    \delta\Tquat{q}_L &= \TqbeEstconj \qprod \Tqbe = \tilde{\Tquat{q}}^* \qprod \Tquat{q} \nonumber \\ 
    &\approx \tilde{\quat{q}}^{*} \qprod \quat{q} + \varepsilon_1 \left(\tilde{\quat{q}}^{'*} \qprod \quat{q} + \tilde{\quat{q}}^{*} \qprod \quat{q}'\right) + \varepsilon_2 \left(\tilde{\quat{q}}^{''*} \qprod \quat{q} + \tilde{\quat{q}}^{*} \qprod \quat{q}''\right)  \nonumber \\ 
    &\approx {\Tquat{I}} + \frac{1}{2} \delta\Tquat{\sigma}_L = {\quat{I}} + \frac{1}{2} \quat{\delta \sigma}_L + \varepsilon_1\frac{1}{2}  \quat{\delta \sigma}^{'}_L + \varepsilon_2\frac{1}{2}  \quat{\delta \sigma}^{''}_L \label{delta_Tq_left}
\end{align}
\normalsize

To simplify the notation, here we denote the true trident quaternion $\Tqbe$ as $\Tquat{q}$, while the estimated quaternion $\TqbeEst$ is referred to as $\TqEst$. We then define the time derivatives as $\dot{\Tquat{q}} = \frac{1}{2} \left(\Tquat{q} \qprod  \Tquat{\omega}_{ib}^b - \Tquat{\omega}_{ie}^e \qprod \Tquat{q}\right)$ and $\dot{\TqEst} = \frac{1}{2} \left(\TqEst \qprod \tilde{\Tquat{\omega}}_{ib}^b - \tilde{\Tquat{\omega}}_{ie}^e \qprod \TqEst\right)$, knowing that $\vect{\tilde{{\omega}}_{ie}^e} = \vect{{{\omega}}_{ie}^e}$, which is the angular velocity of the Earth. 

In expression \eqref{delta_Tq_left}, we define $\delta\Tquat{\sigma}_{L} = \quat{\delta \sigma}_{L} + \varepsilon_1 \quat{\delta \sigma}_{L}' + \varepsilon_2 \quat{\delta \sigma}_{L}''$ such that $ \quat{\delta \sigma}_{L}, \quat{\delta \sigma}_{L}', \quat{\delta \sigma}_{L}''$ are quaternion vectors, implying that $\quat{\delta \sigma}_{L}=(\vect{\delta{\sigma}_{L}},0), \quat{\delta \sigma}_{L}'=(\vect{\delta{\sigma}_{L}'},0), \quat{\delta \sigma}_{L}''=(\vect{\delta{\sigma}_{L}''},0)$.

\begin{equation}
   \quat{\delta q}_{L} = \tilde{\quat{q}}^{*} \qprod \quat{q}  \approx {\quat{I}} + \frac{1}{2} \quat{\delta \sigma}_{L} = {\quat{I}} + \frac{1}{2}\quat{\phi}^{b} 
\end{equation}

From this result, it follows that
$\quat{\delta \sigma}_{L} = \quat{\phi}^{b} \Rightarrow   \vect{\delta{\sigma}_{L}} = \vect{{\phi}^{b}}$.
\begin{align}
   \quat{\delta \sigma}_{L}' &= \tilde{\quat{q}}^{*} \qprod (\quat{v}^e - \tilde{\quat{v}}^e) \qprod \quat{q} = -\tilde{\quat{q}}^{*} \qprod \delta \quat{v}^e \qprod \quat{q} \nonumber \\
   &= -\tilde{\quat{q}}^{*} \qprod \quat{q} \qprod \quat{q}^* \qprod \delta \quat{v}^e \qprod \quat{q} \nonumber \\ 
   &\approx -({\quat{I}} + \frac{1}{2} \quat{\delta \sigma}_{L}) \qprod \quat{q}^* \qprod \delta \quat{v}^e \qprod \quat{q} \approx -\quat{q}^* \qprod \delta \quat{v}^e \qprod \quat{q} \nonumber \\ 
   &= (-\matr{C_e^b} \delta \vect{v}^e,0) \label{delta_1}
\end{align}
\begin{align}
   \quat{\delta \sigma}_{L}'' &= \tilde{\quat{q}}^{*} \qprod (\quat{p}^e - \tilde{\quat{p}}^e) \qprod \quat{q} = -\tilde{\quat{q}}^{*} \qprod \delta \quat{p}^e \qprod \quat{q}  \nonumber \\
   &= -\tilde{\quat{q}}^{*} \qprod \quat{q} \qprod \quat{q}^* \qprod \delta \quat{p}^e \qprod \quat{q} \nonumber \\ 
   &\approx -({\quat{I}} + \frac{1}{2} \quat{\delta \sigma}_{L}) \qprod \quat{q}^* \qprod \delta \quat{p}^e \qprod \quat{q} \approx - \quat{q}^* \qprod \delta \quat{p}^e \qprod \quat{q} \nonumber \\ 
   &= (-\matr{C_e^b}\delta \vect{p}^e,0) \label{delta_2}
\end{align}

In \eqref{delta_1} and \eqref{delta_2}, second-order terms are considered negligible,
then:
\begin{equation}
    \delta \dot{\Tquat{q}}_L \approx \frac{1}{2} \delta\dot{\Tquat{\sigma}}_{L} = \frac{1}{2} (\delta\dot{\quat{\sigma}}_{L} + \varepsilon_{1} \delta\dot{\quat{\sigma}}_{L}'  + \varepsilon_{2} \delta\dot{\quat{\sigma}}_{L}'' )
\end{equation}
From \eqref{eq_phib} we know that:
\begin{equation}
\vect{\delta\dot{\sigma}_{L}} = -(\wibb)_\times \vect{\delta \sigma}_{L} - \Dwibb = \vect{\dot{\phi}^b}, \label{dsigma_punto}   
\end{equation}
\noindent in the same manner, the dynamics of $\vect{\delta\dot{{\sigma}}_{L}'}$ and $\vect{\delta\dot{{\sigma}}_{L}''}$ are obtained:
\begin{align}
    \vect{\delta\dot{{\sigma}}_{L}'} &= -\dot{\matr{C_e^b}} \vect{\delta v^e}- \matr{C_e^b} \dot{\vect{\delta v^e}}
    = S(\vect{f^b}) \vect{\delta{\sigma}_{L}} - \vect{\delta f^b} + \nonumber \\
    &(-S(\wibb) -  S(\matr{C_e^b}\vect{\omega_{ie}^e})) \delta{\sigma}_{L}' +\vect{\delta \gamma^e} 
    \label{dsigma_punto_1}
\end{align}
\begin{align}
    \vect{\delta\dot{{\sigma}}_{L}''} &= -\dot{\matr{C_e^b}} \vect{\delta p^e} - \matr{C_e^b} \vect{\delta v^e} \nonumber \\ 
    &=  \vect{\delta{\sigma}_{L}'} + (-S(\wibb) + S(\matr{C_e^b}\vect{\omega_{ie}^e}) )\vect{\delta{\sigma}_{L}''}, \label{dsigma_punto_2}
\end{align}
\noindent where:
\begin{equation}
\vect{\delta \gamma^e} = \frac{\partial \vect{\gamma^e}}{\partial \vect{p^e}} \bigg|_{\vect{p^e}} \delta \vect{p^e} = -\frac{\partial \vect{\gamma^e}}{\partial \vect{p^e}} \bigg|_{\vect{p^e}} \matr{C_b^e} \vect{\delta{\sigma}_{L}''}.    
\end{equation}

Knowing the equations of $\vect{\delta\dot{{\sigma}}_{L}}$, $\vect{\delta\dot{{\sigma}}_{L}'}$ and $\vect{\delta\dot{{\sigma}}_{L}''}$, based on equations \eqref{eq_Dfb}, the differential error model of the trident quaternion can be written in its matrix form and linearized to the left-hand side as:

\begin{equation}
    \vect{\delta \dot{x}_L}= \matr{F_L} \vect{\delta x_L} + \matr{B_L} \vect{w}, \text{ where }\label{eq:errorQT}
\end{equation}
\begin{equation}
    \vect{\delta x_L} = \begin{bmatrix}
        \vect{\delta{{\sigma}}_{L}} & \vect{\delta{{\sigma}}_{L}'} & \vect{\delta{{\sigma}}_{L}''} & \vect{\delta b_g} & \vect{\delta b_a}     \end{bmatrix} ^{T}
\end{equation}
\begin{equation}
     \vect{w} = \begin{bmatrix}
        \vect{\xi_{{\omega}}} & \vect{\xi_{{f}}} & \vect{\xi_{b_{\omega}}} & \vect{\xi_{b_{f}}}   \end{bmatrix} ^{T}
\end{equation}
\begin{equation*}
\matr{F_L} =  \begin{bmatrix}
    -(\wibb)_\times & \0_{3 \times 3} & \0_{3 \times 3} & -\I_{3 \times 3} & \0_{3 \times 3} \\
    (\vect{f^b})_\times & -(\wibb + \matr{C_e^b}\vect{\omega_{ie}^e})_\times & \matr{C_e^b} \left( \frac{\partial \vect{\gamma^e}}{\partial \vect{p^e}} \right)_{(\vect{p^e})} \matr{C_e^b} & \0_{3 \times 3} & -\I_{3 \times 3} \\
    \0_{3 \times 3} & \I_{3 \times 3} & -(\wibb - \matr{C_e^b}\vect{\omega_{ie}^e})_\times & \0_{3 \times 3} & \0_{3 \times 3} \\
    \0_{3 \times 3} & \0_{3 \times 3} & \0_{3 \times 3} & \0_{3 \times 3} & \0_{3 \times 3} \\
    \0_{3 \times 3} & \0_{3 \times 3} & \0_{3 \times 3} & \0_{3 \times 3} & \0_{3 \times 3}
\end{bmatrix}, 
\end{equation*}

\small
\begin{equation}
B_L = \begin{bmatrix}
    -I_{3 \times 3} & \0_{3 \times 3} & \0_{3 \times 3} & \0_{3 \times 3} \\
    \0_{3 \times 3} & -I_{3 \times 3} & \0_{3 \times 3} & \0_{3 \times 3} \\
    \0_{3 \times 3} & \0_{3 \times 3} & \0_{3 \times 3} & \0_{3 \times 3} \\
    \0_{3 \times 3} & \0_{3 \times 3} & \I_{3 \times 3} & \0_{3 \times 3} \\
    \0_{3 \times 3} & \0_{3 \times 3} & \0_{3 \times 3} & \I_{3 \times 3}
\end{bmatrix}
\end{equation}
\normalsize

It can be seen that the relationship between the trident quaternion error and equation \eqref{eq:INSerror}, $\vect{\delta x_L}$ and $\vect{\delta x}$ is given by $\vect{\delta x_L} = \matr{J_L} \vect{\delta x} \Leftrightarrow \vect{\delta x} = \matr{J_L}^T \vect{\delta x_L}$, where:
\small
\begin{equation}
    \matr{J_L} = \begin{bmatrix}
    \matr{C_b^e} & \0_{3 \times 3} & \0_{3 \times 3} & \0_{3 \times 3} & \0_{3 \times 3} \\
    \0_{3 \times 3} & -\matr{C_b^e} & \0_{3 \times 3} & \0_{3 \times 3} & \0_{3 \times 3} \\
    \0_{3 \times 3} & \0_{3 \times 3} & -\matr{C_b^e} & \0_{3 \times 3} & \0_{3 \times 3} \\
    \0_{3 \times 3} & \0_{3 \times 3} & \0_{3 \times 3} & I_{3 \times 3} & \0_{3 \times 3} \\
    \0_{3 \times 3} & \0_{3 \times 3} & \0_{3 \times 3} & \0_{3 \times 3} & I_{3 \times 3}
\end{bmatrix} \label{eq:J_L}
\end{equation}
\normalsize

The left-hand side error covariance matrix of the system, defined as $\matr{P_{\delta x_L}} = E[\vect{\delta x_{L}} \vect{\delta x_{L}}^T]$, is related to equation \eqref{eq:INSerror}, through $\matr{P_{\delta x_L}}=\matr{J_L}\matr{P_{\delta x}}\matr{J_L}^T.$





Let $\vect{\tilde{p}^e_k}$ and $\vect{\tilde{v}^e_k}$ be the position and velocity measurements, respectively (in the $e$ frame), provided by the GPS at time instant $k$. The model of these measurements can be expressed as: $\vect{{p}^e_k} = \vect{\tilde{p}^e_k + n_{p_k}}$, $\vect{{v}^e_k} = \vect{\tilde{v}^e_k + n_{v_k}}$. 
Where the vector $\begin{bmatrix} \vect{\eta_{p_k}} & \vect{\eta_{v_k}} \end{bmatrix}^T$ corresponds to the GPS measurement noise (assumed to be white Gaussian with zero mean and known variance), while $\vect{{p}^e_k}$ and $\vect{{v}^e_k}$ are the true magnitudes. 
The model of the measurements associated with the left-hand side error model of the trident quaternion is:

\begin{equation}
\vect{\delta y_k} = 
\begin{bmatrix}
\vect{\tilde{p}^e_k} - \vect{p_k^e} \\
\vect{\tilde{v}^e_k} - \vect{v_k^e}
\end{bmatrix} + \begin{bmatrix}
\vect{\eta_{p_k}} \\
\vect{\eta_{v_k}}
\end{bmatrix} = \matr{H_L} \vect{\delta x_{L_k}} + \begin{bmatrix}
\vect{\eta_{p_k}} \\
\vect{\eta_{v_k}}
\end{bmatrix} 
\end{equation}

\noindent with 
\begin{align}
\matr{H_L} = \begin{bmatrix}
    \0_{3 \times 3} & \0_{3 \times 3} & -\matr{C_b^e} & \0_{3 \times 3} & \0_{3 \times 3} \\
    \0_{3 \times 3} & -\matr{C_b^e} & \0_{3 \times 3} & \0_{3 \times 3} & \0_{3 \times 3}
\end{bmatrix} 
\end{align}

Given this information and the state model in equation \eqref{eq:errorQT}, along with the hypotheses about the sensor noises, it is possible to apply an EKF to estimate the left-hand side error of the trident quaternion $\vect{\delta x_L}$, thereby compensating for the error obtained by the inertial navigator.

\section{Results}
\label{sec:Results}

\begin{figure}[t]
    \centering
    \includegraphics[width=0.75\linewidth]{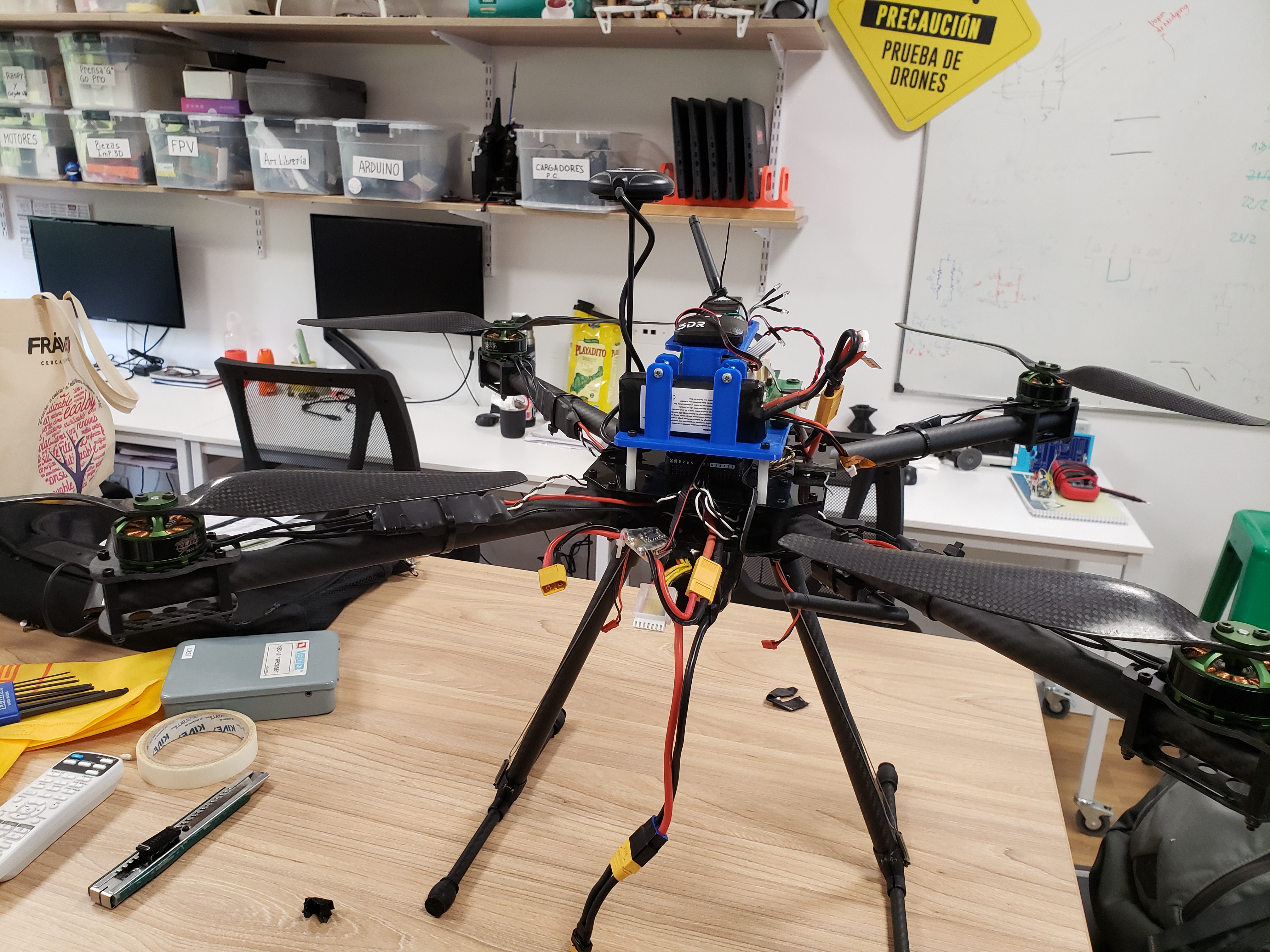}
    \caption{Multi-rotor UAV used for experimental testing.}
    \label{fig:cuad}
\end{figure}

To experimentally validate the proposed algorithm, a symmetric X-configured multi-rotor UAV developed in our lab was used (Figure \ref{fig:cuad}). The platform consists of a carbon fiber frame with a maximum motor separation of 70 cm. It is equipped with Multistar 3508 268KV motors, T-Motor 14''×4.8'' carbon fiber propellers, Hobbyking 40A ESCs, and a 6S 5000mAh battery. The vehicle's main autopilot is a Pixhawk 2.4.8 with an integrated GPS and magnetometer, running PX4 firmware. In addition to controlling the vehicle, the Pixhawk logs navigation data to serve as a reference for evaluating an alternative autopilot implementing the navigation algorithm based on trident quaternions.

A custom-developed board was also mounted on the multi-rotor UAV. This board features a Cortex-M7 microcontroller with a double-precision floating-point unit, an MPU9250 IMU, and a GPS receiver using UBlox NEO-M7 messages via a serial port, providing position and velocity data at 1 Hz. The quaternion-based navigation algorithm was implemented in C and executed on this board under a FreeRTOS environment. Since no precise alignment was performed between the custom board and the Pixhawk, a bias in the pitch and roll estimates between both systems was expected.

The comparison was not intended to be exhaustive; rather, the goal was to assess whether there was a general correspondence between the navigation data provided by each system. A more detailed analysis is planned for future work, which would require a higher-quality sensor than the Pixhawk used in this study. Additionally, to ensure a fair comparison, both systems should rely on the same type of sensors. While our algorithm uses only inertial and GPS data, the Pixhawk autopilot also incorporates an altimeter and a magnetometer, introducing potential discrepancies.

The primary objective here was to evaluate the execution of the trident-quaternion-based navigation code on our autopilot and verify the coherence of the obtained data.

During the flight experiments, measures were taken to mitigate vibrations and ensure stable data acquisition. The IMU was mounted on shock absorbers to reduce the impact of external disturbances. Additionally, IMU data was recorded at a sampling rate of 200 Hz, with a digital low-pass filter set to 20 Hz.

Several flights were conducted with the multi-rotor UAV under various conditions, including hover, maneuvers, and altitude changes. This provided a comprehensive data set to evaluate the performance and effectiveness of the algorithm in comparison with the Pixhawk results. Below are the results from some of the tests conducted.

\begin{figure}[t]
    \centering
    \includegraphics[width=0.98\linewidth]{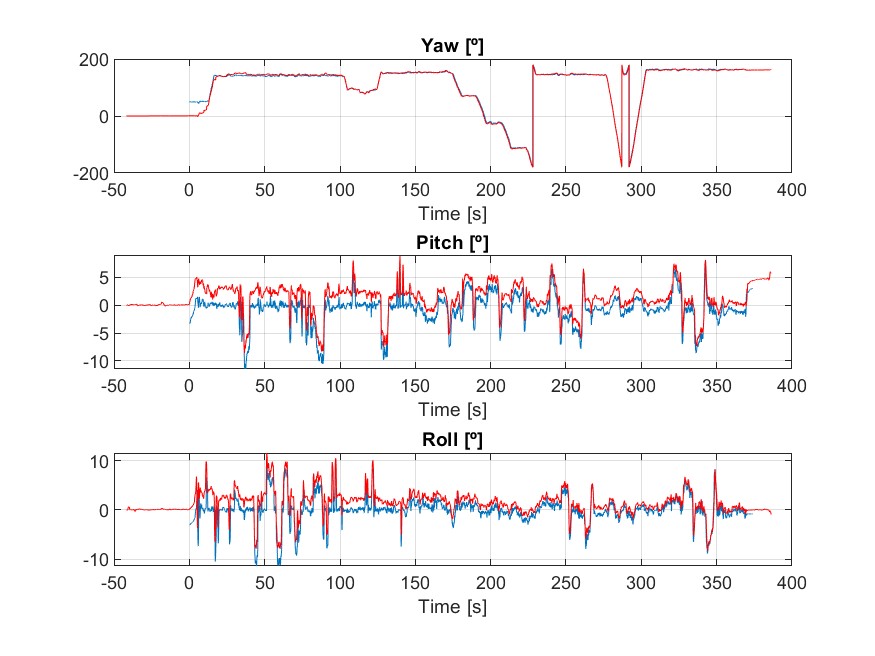}
    \caption{Estimation of the vehicle orientation (Blue: PixHawk, Red: Trident quaternions).}
    \label{fig:results_ypr}
\end{figure}

In Figure \ref{fig:results_ypr}, the attitude estimation provided by the Pixhawk and our algorithm is shown. Specifically, it can be seen that the yaw estimation from our board is incorrect at the beginning of the navigation. This is because, unlike the Pixhawk, our board does not use the magnetometer. Therefore, the vehicle's trajectory must sufficiently excite the system modes for yaw to become observable and allow for correct estimation. It is observed that when the vehicle starts moving, the yaw estimation corrects itself and properly follows the Pixhawk's estimation for the remainder of the trajectory. While the magnetometer could have been included as an external sensor, it does not seem necessary for an application like this, unless the vehicle remains static for extended periods, making yaw unobservable. However, this sensor will likely be included in future tests.

\begin{figure}[t]
    \centering
    \includegraphics[width=0.98\linewidth]{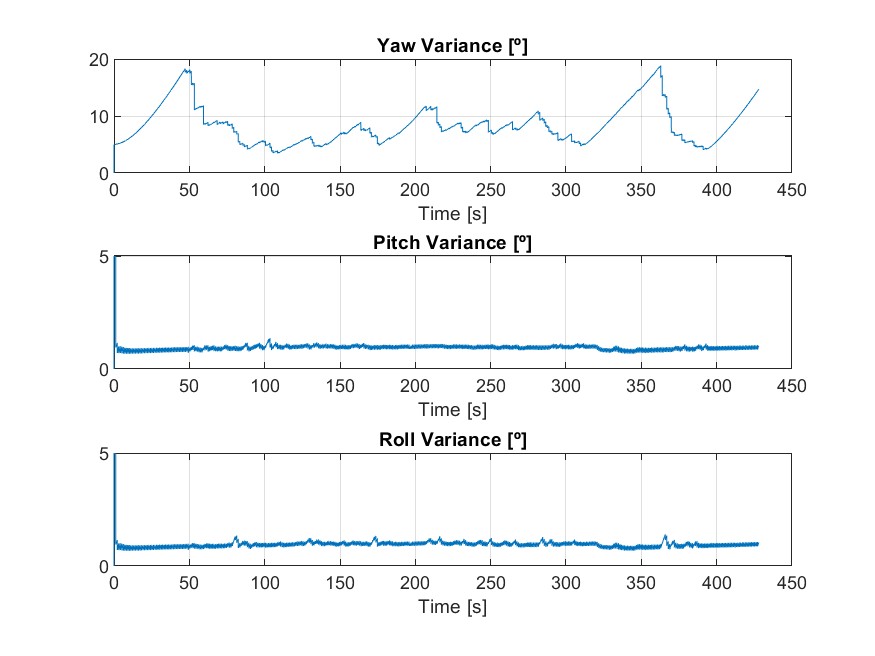}
    \caption{Estimation of the attitude covariance.}
    \label{fig:AttitudeCov}
\end{figure}

Regarding roll and pitch, a bias can be observed between the estimations, which is due to the lack of alignment between the boards. The difference between the two boards appears to decrease as the estimation uncertainty is reduced. Figure \ref{fig:AttitudeCov} shows the attitude covariance estimated by the algorithm. 

\begin{figure}[t]
    \centering
    \includegraphics[width=0.98\linewidth]{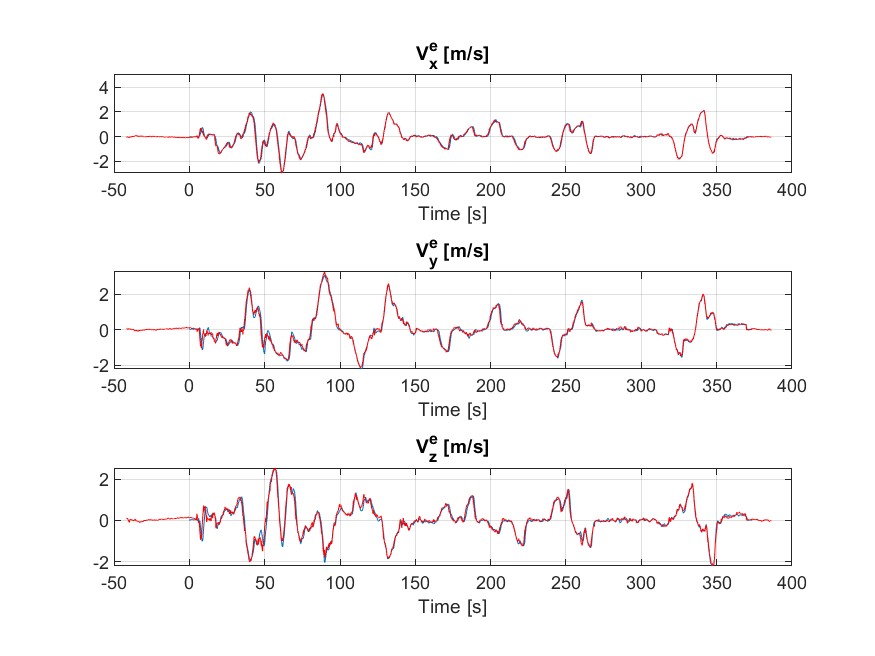}
    \caption{Estimation of the vehicle velocity in the ECEF frame (Blue: PixHawk, Red: Trident quaternions).}
    \label{fig:results_vel}
\end{figure}

On the other hand, in Figure \ref{fig:results_vel}, it is observed that the velocity estimation is consistent with the data provided by the Pixhawk (in blue) throughout the trajectory. Working with trident quaternions is advantageous, as they encapsulate velocity information along with position and attitude data. This approach of using a more general scheme can even be applied to problems such as spacecraft dynamics representation \cite{Stanfield2023} and coordinated vehicle control, where dual quaternions have already demonstrated their usefulness, as well as their limitations \cite{Giribet, Marciano2024, Farias2024}.

\begin{figure}[t]
    \centering
    \includegraphics[width=0.98\linewidth]{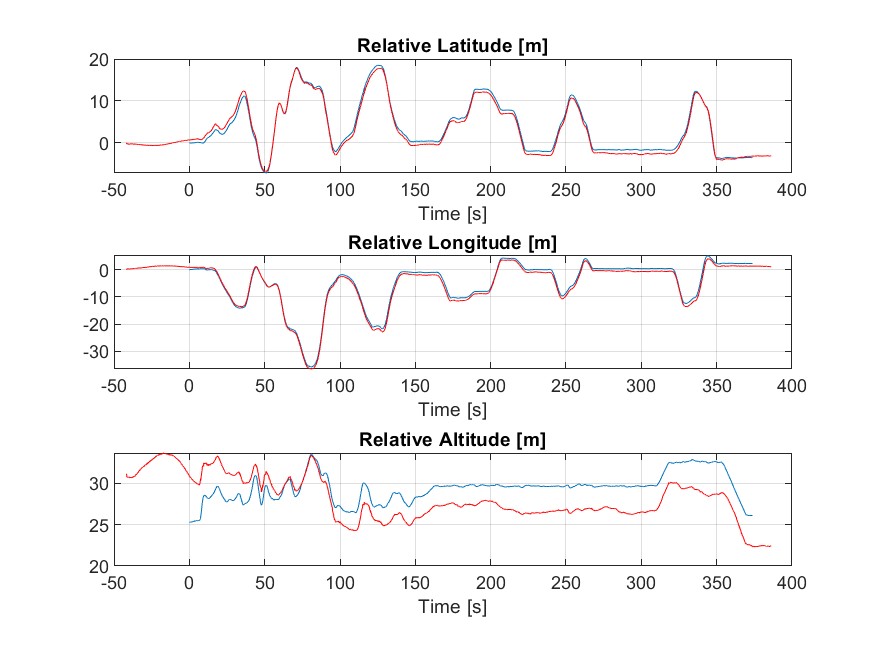}
    \caption{Estimation of the vehicle position (Blue: PixHawk, Red: Trident quaternions).}
    \label{fig:results_lla_metros}
\end{figure}

\begin{figure}[t]
    \centering
    \includegraphics[width=0.98\linewidth]{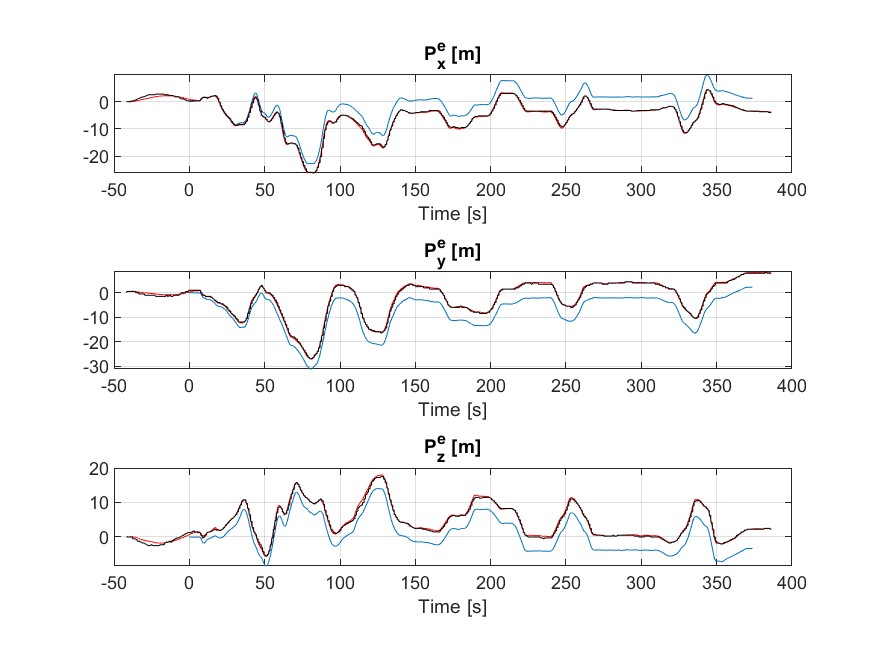}
    \caption{Estimation of the vehicle position in the ECEF frame (Blue: PixHawk, Red: Trident quaternions, Black: GPS).}
    \label{fig:results_pos}
\end{figure}

In Figure \ref{fig:results_lla_metros}, a bias in altitude is observed compared to Pixhawk's estimation. This is because the Pixhawk integrates GPS data and also uses the barometric altimeter for altitude estimation. The altitude bias is not due to the algorithm but to the GPS measurement itself. This discrepancy can also be observed in Figure \ref{fig:results_pos}, where the estimation provided by our algorithm (in red) is close to the measurement provided by the GPS sensor (in black). The integration of a barometric altimeter measurement into our algorithm is proposed as future work.

\begin{figure}[!htb]
    \centering
    \includegraphics[width=0.98\linewidth]{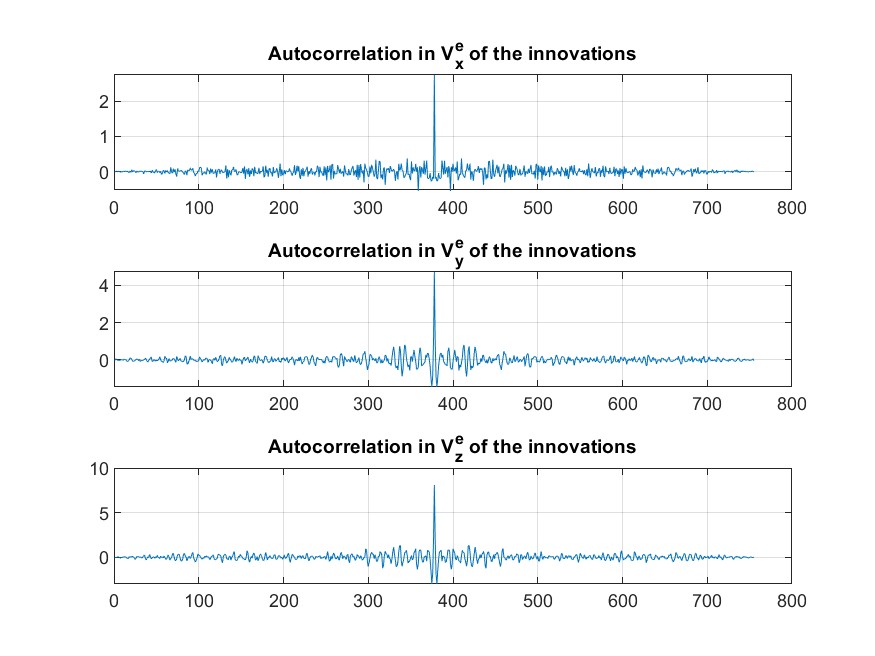}
    \caption{Autocorrelation of velocity innovations in the ECEF frame}
    \label{fig:inn_corr_ve_v1}
\end{figure}

\begin{figure}[!htb]
    \centering
    \includegraphics[width=0.98\linewidth]{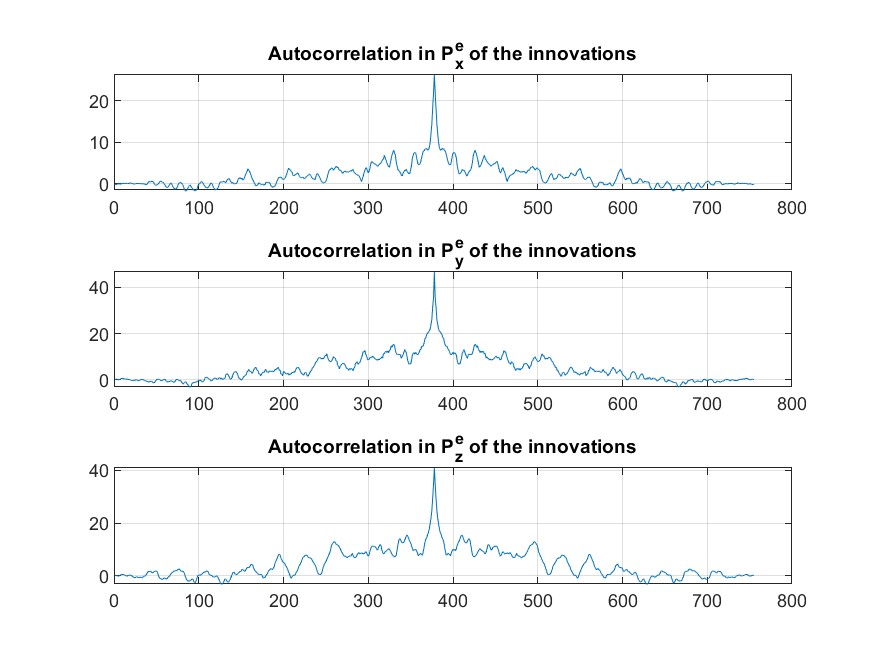}
    \caption{Autocorrelation of position innovations in the ECEF frame}
    \label{fig:inn_corr_pe_v1}
\end{figure}

To verify the behavior of the algorithm, the innovation process was recorded, and the autocorrelation was computed. This can be observed in Figures \ref{fig:inn_corr_ve_v1} and \ref{fig:inn_corr_pe_v1} for velocity and position, respectively. Although, in theory, the autocorrelation should be a delta function, this is not the case here, which is expected in a loosely coupled implementation of an integrated navigation algorithm.



\begin{figure}[!htb]
    \centering
    \includegraphics[width=0.98\linewidth]{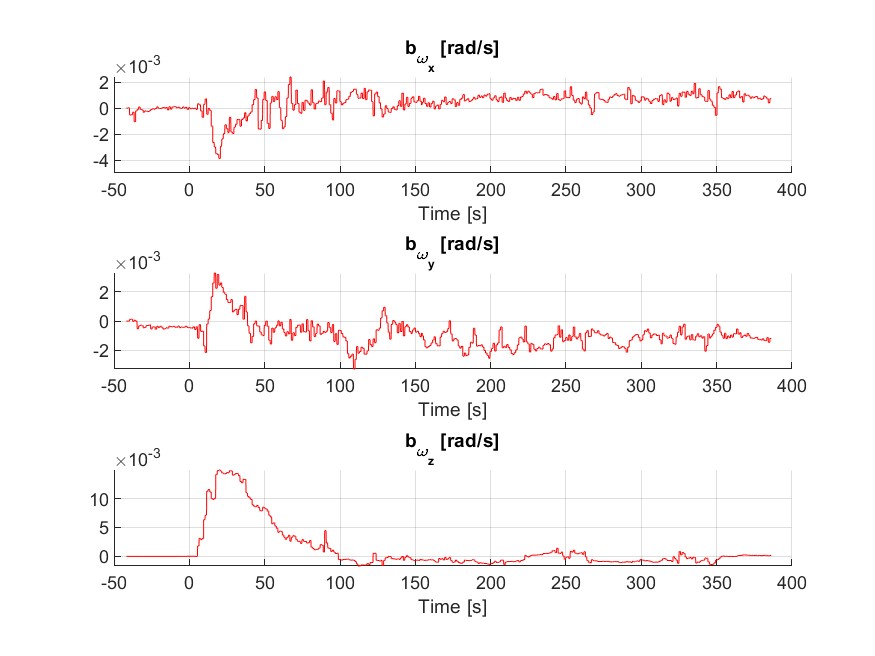}
    \caption{Gyroscope Bias Parameters}
    \label{fig:sesgo_w_v1}
\end{figure}

\begin{figure}[!htb]
    \centering
    \includegraphics[width=0.98\linewidth]{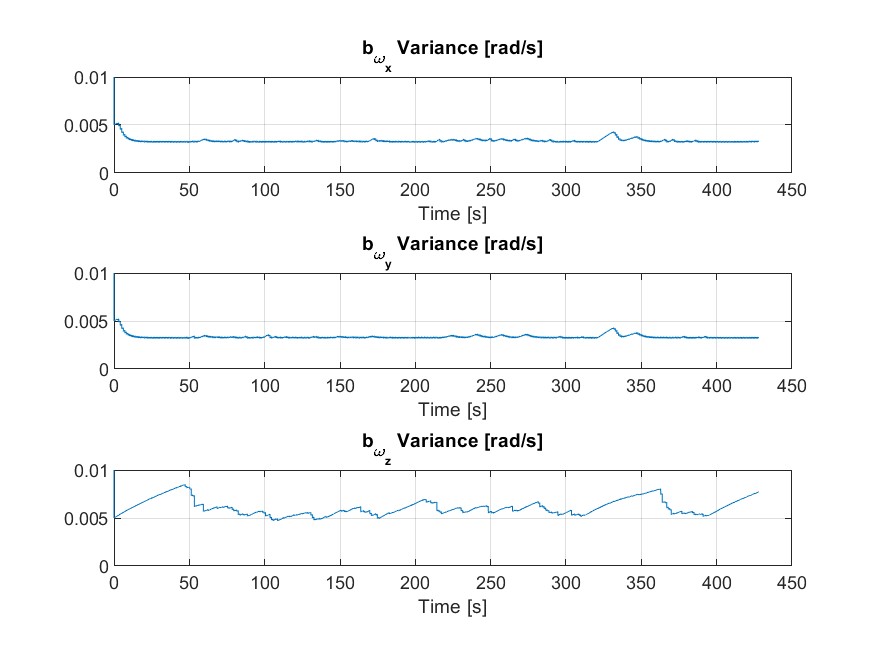}
    \caption{Variance of Gyroscope Bias Instabilities}
    \label{fig:var_bw_v1}
\end{figure}

\begin{figure}[!htb]
    \centering
    \includegraphics[width=0.98\linewidth]{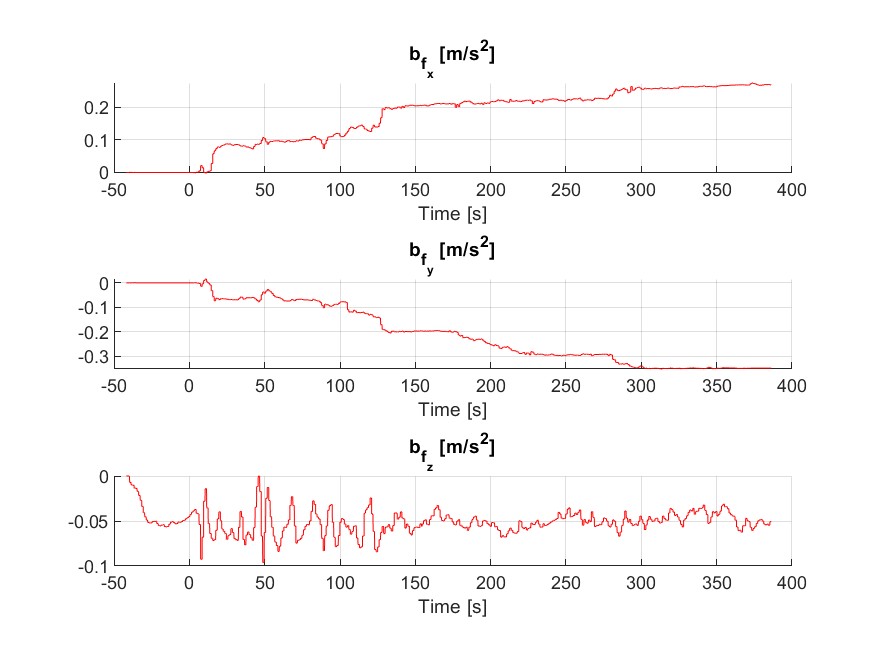}
    \caption{Accelerometer Bias Parameters}
    \label{fig:sesgo_f_v1}
\end{figure}

\begin{figure}[!htb]
    \centering
    \includegraphics[width=0.98\linewidth]{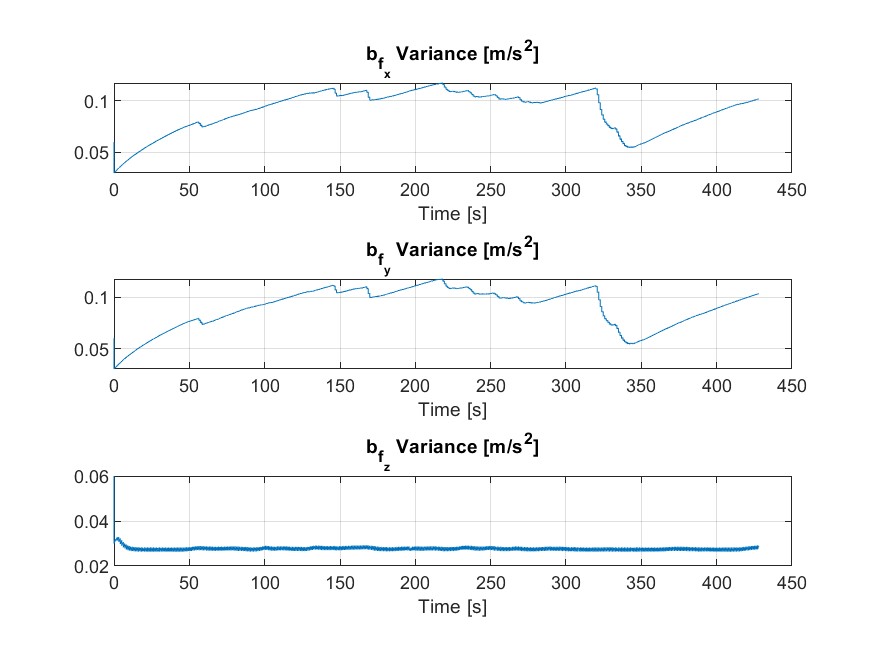}
    \caption{Variance of Accelerometer Bias Instabilities}
    \label{fig:var_bf_v1}
\end{figure}

Regarding the estimation of the inertial sensor error parameters, Figures \ref{fig:sesgo_w_v1}, \ref{fig:var_bw_v1}, \ref{fig:sesgo_f_v1} and \ref{fig:var_bf_v1} show the estimated biases of the gyroscopes and accelerometers, respectively, along with the corresponding covariances.

\section{Conclusions}
\label{sec:Conclusions}


The findings of this study reinforce the idea that the incorporation of trident quaternions provides a compact and unified mathematical framework for representing navigation equations. By encapsulating multiple sensor measurements within a single mathematical object, this approach enhances the accuracy and robustness of navigation systems. From a numerical standpoint, trident quaternions share the computational advantages of conventional quaternions while also introducing benefits in terms of implementation efficiency. Additionally, their structured algebraic properties open new possibilities for designing inertial navigation algorithms and observer schemes tailored to specific applications.

The experimental results demonstrate a strong consistency between the navigation data obtained from the proposed algorithm and the reference Pixhawk autopilot. Notably, the yaw estimation error observed at the beginning of the trajectory is a direct consequence of the absence of a magnetometer in the proposed system, highlighting the importance of trajectory excitation for observability. Despite the expected biases in roll and pitch due to the lack of precise alignment between the custom board and the Pixhawk, the estimation trends align well, with the discrepancies diminishing as uncertainty is reduced. Furthermore, velocity estimates closely match those of the Pixhawk, demonstrating the feasibility of using trident quaternions to encapsulate position, attitude, and velocity information within a single framework.

Future work will focus on enhancing the system by integrating additional sensors such as barometric altimeters, magnetometers, or vision-based systems to further refine navigation accuracy. The implementation of a control algorithm within this formalism is also planned, leveraging the advantages of trident quaternions to improve trajectory tracking and system stability. Moreover, exploring alternative sensor fusion strategies and extending the methodology to other applications, such as coordinated multi-vehicle control, could further demonstrate the potential of this approach in real-world scenarios.

\bibliographystyle{IEEEtran}
\bibliography{cluster}

\vspace{12pt}

\end{document}